\documentclass[
]{ceurart}

\usepackage{graphicx}
\usepackage{geometry}
\usepackage{caption}
\usepackage{subcaption}
\begin{document}

\copyrightyear{2024}
\copyrightclause{Copyright for this paper by its authors.
  Use permitted under Creative Commons License Attribution 4.0
  International (CC BY 4.0).}

\conference{AI for Access to Justice Workshop at JURIX 2024}

\title{Analyzing Images of Legal Documents: Toward Multi-Modal LLMs for Access to Justice}

%Toward filling in forms based on images using LLMs
%From pictures of paper forms to structured data: Toward multi-modal LLMs for access to justice

% \tnotetext[1]{You can use this document as the template for preparing your
%   publication. We recommend using the latest version of the ceurart style.}

\author[1]{Hannes Westermann}[%
orcid=0000-0002-4527-7316,
email=hannes.westermann@maastrichtuniversity.nl,
]
\cormark[1]
\address[1]{Maastricht Law and Tech Lab, Maastricht University, Maastricht, Netherlands}

\author[2]{Jaromir Savelka}[%
orcid=0000-0002-3674-5456,
email=jsavelka@cs.cmu.edu,
url=https://www.cs.cmu.edu/~jsavelka/,
]
\address[2]{Computer Science Department, Carnegie Mellon University, Pittsburgh, USA}

\cortext[1]{Corresponding author.}
% \fntext[1]{These authors contributed equally.}

\begin{abstract}
   Interacting with the legal system and the government requires the assembly and analysis of various pieces of information that can be spread across different (paper) documents, such as forms, certificates and contracts (e.g. leases). This information is required in order to understand one's legal rights, as well as to fill out forms to file claims in court or obtain government benefits. However, finding the right information, locating the correct forms and filling them out can be challenging for laypeople. Large language models (LLMs) have emerged as a powerful technology that has the potential to address this gap, but still rely on the user to provide the correct information, which may be challenging and error-prone if the information is only available in complex paper documents. We present an investigation into utilizing multi-modal LLMs to analyze images of handwritten paper forms, in order to automatically extract relevant information in a structured format. Our initial results are promising, but reveal some limitations (e.g., when the image quality is low). Our work demonstrates the potential of integrating multi-modal LLMs to support laypeople and self-represented litigants in finding and assembling relevant information.
   
  %Identifying and synthesizing the information contained in these documents is 
  
  %Individuals are often required to fill in forms in order to get access to vital governmental services as well as to fulfill various legal obligations. Filling in forms correctly may often represent a significant hurdle for laypeople, as they may find it difficult to understand which forms to use and which elements need to be present. Hence, their interaction with the justice system and government can be impaired as they might be prevented from properly enforcing their legal rights, obtaining government benefits, or engaging with various procedures in efficient ways. 
  %Large language models (LLMs) have emerged as a powerful technology that has the potential to address this gap. Most of the existing work has focused on using LLMs to assist in filling the forms or extracting relevant information from their textual representation. However, it is quite common for forms to be printed on paper and filled in by handwriting. We present an investigation into utilizing mutli-modal LLMs to analyze images of forms originally in paper format in order to automatically extract relevant information they contain. Our initial results are promising, but reveal several limitations that are present in case the image quality is low. Hence, our work demonstrates the potential of integrating multi-modal generative LLMs in simplifying many aspects of processing forms correctly. 
\end{abstract}


\maketitle

\section{Introduction}
Many people lack adequate access to justice, i.e., they are not able to resolve their legal problems \cite{currie2009legal}, which can result in significant costs to society \cite{semple2015cost}. In addition, many individuals cannot take full advantage of the benefits provided by the government. The Justice Task Force estimates that 1.5 billion people have justice problems they cannot resolve, while 4.5 billion people are excluded from the opportunities the law provides. \cite{Justice_For_All}.

One hurdle to overcoming the issue of access to justice lies in the difficulty of laypeople in filling out forms and drafting legal documents. For example, self-represented litigants struggle with deciding which forms to use and how to fill them out \cite{macfarlane2013national}. Further, researchers have found that laypeople often struggle with understanding how to craft a legally sound narrative (e.g., by failing to understand which facts to establish, and actually establishing them) \cite{branting2020judges}.
%Forms can thus present a significant hurdle for obtaining effective legal relief and interacting with the court system, and may cause people to struggle to obtain governmental benefits or licenses. 
Sunstein refers to the issue of burdening people as ``administrative sludge'' and estimates that almost 10 billion hours of paperwork were imposed on people in 2015 in the United States alone \cite{sunstein2018sludge}.

Artificial intelligence (AI) has emerged as a powerful way to support users in filling out forms and supporting them in understanding their rights. AI systems can ask questions or analyze fact descriptions to provide people with information about their legal rights \cite{westermann_justicebot_2023,westermann2023bridging}, or guide them through the process of filling out forms \cite{steenhuis2023weaving}. Large language models (LLMs) can even synthesize responses based on the information provided \cite{westermanndallma}.

AI, and LLMs specifically, power cutting-edge methods to bridge the gap between lay persons' knowledge and legally relevant formulations as used in, e.g., forms or court submissions. However, they often assume that the user can provide the relevant information for processing on a computer system. This may introduce an element of friction, since much of the information may be contained in paper forms, certificates, contracts, letters or other documents. The user then has to find the relevant information and type it into the computer which may be difficult for some people. Further, the required information may itself be contained in complicated forms or documents, making it tricky for the user to determine which of the information is the one that is needed by the AI system. 

% Interacting with the legal system and the government requires the assembly and analysis of various pieces of information that can be spread across different documents, such as legislation and case law, but also individual (paper) documents such as forms, contracts, letters and certificates. These documents contain important information that is required for e.g. understanding ones legal rights, filling in forms, and fulfilling procedures.
% The difficulty in filling in forms can stem from several causes. RateMyPDF discuss the difficulties of different types of fields in forms.

In this paper, we conduct an initial investigation as to whether multi-modal LLMs are able to reliably extract information from photographs (i.e., images) of forms. This could enable AI systems to simply instruct the user to take images of relevant documents, and then automatically identify and extract the needed information, for use in, e.g., filling out other forms, informing users of their rights and legal situation, or suggesting drafts of court submissions.

To explore the described capability of LLMs, we investigate the following research questions:

\begin{enumerate}
    \item[RQ1] To what extent are multi-modal LLMs (GPT-4o) able to reliably extract structured data from photographs of forms?
    \item[RQ2] How do factors such as image quality, complex data, missing fields and unclear handwriting affect the results?
\end{enumerate}

\section{Related Work}

\subsubsection*{AI for Access to Justice}
Access to justice initiatives have evolved significantly from early expert systems to modern AI-powered solutions. Early work by \cite{paquin1991loge} demonstrated the complexity of formalizing legal knowledge, showing how even seemingly straightforward legal questions required understanding of both broad legal principles and granular details. This challenge became increasingly important as courts faced growing numbers of self-represented litigants, leading to innovations like the Protection Order Advisory system \cite{branting2001advisory}, which demonstrated how inference and document-drafting techniques could assist pro se litigants in cases where prima facie requirements did not demand complex legal reasoning. Further advances came through web-based legal decision support systems \cite{zeleznikow2002using}, which not only provided support for unrepresented litigants but also facilitated dispute negotiation in domains such as Family Law and legal aid eligibility. More recently, sophisticated approaches like the JusticeBot methodology \cite{westermann_justicebot_2023,Westermann_thesis_2023} have emerged, combining case-based and rule-based reasoning to help laypeople explore their legal rights, demonstrating success in real-world applications such as landlord-tenant disputes where thousands of users have benefited from automated legal information and guidance.

\subsubsection*{LLMs for Extracting Information from Legal Texts}
Recent technological advances have enabled more sophisticated approaches to extracting information from legal texts. For instance, \cite{westermann2023bridgingn} developed a system that automatically maps laypeople's factual descriptions to relevant legal issues with 93.5\% accuracy, helping users identify potential legal remedies. This work is complemented by tools like RateMyPDF \cite{steenhuis2023beyond}, which helps improve the usability of court forms through automated analysis and suggestions. The field continues to advance with innovative approaches such as narrative-driven case elicitation \cite{branting2023narrative}, which uses schemas induced from legal case corpora to distinguish relevant facts and identify critical distinctions between competing legal hypotheses. In this paper, we expand on this important line of work by introducing the image modality into the task of extracting information. Instead of requiring the user to manually enter the relevant information, this could open the door to systems that merely require the user to snap an image of their documents, allowing the system to automatically locate, retrieve, and process the information.

\subsubsection*{LLMs for Legal Tasks and Access to Justice}
LLMs have emerged as powerful tools in the legal domain, with numerous studies investigating their capabilities across various legal tasks. Models such as OpenAI's GPT-4 \cite{openai2023gpt4} have demonstrated remarkable versatility in handling diverse legal challenges. Researchers have extensively explored LLMs' abilities in several key areas: legal reasoning, document analysis and annotation, and access to justice applications. In terms of legal reasoning, studies have shown that LLMs can effectively handle complex legal tasks, such as answering legal entailment questions in the Japanese bar exam \cite{https://doi.org/10.48550/arxiv.2212.01326}, passing the Uniform Bar Examination \cite{katz2023gpt,katz2024gpt}, and conducting statutory reasoning \cite{blair2023can,nguyen2023blackbox}. Additionally, LLMs have been successfully applied to modeling US Supreme Court opinions \cite{hamilton2023blind}. In the realm of document analysis, LLMs have proven valuable for annotating court opinions \cite{savelka2023can,gray2024empirical} as well as other legal documents \cite{savelka2023unlocking,savelka2023unreasonable} and explaining the meaning of statutory terms \cite{savelka2023explaining}. They have also shown promise in dispute mediation \cite{westermann2023llmediator,tan2024robotsmiddleevaluatingllms} and providing legal information to laypeople \cite{tan2023chatgpt}. Particularly significant is the growing role of LLMs in improving access to justice. Recent research has demonstrated their capability to transform legal articles into structured representations for legal decision support tools \cite{janatian2023text} and create automated forms \cite{steenhuis2023weaving}, as well as facilitating the assessment of eligibility for legal aid \cite{steenhuis2024gettingdoorstreamliningintake}. LLMs have also been successfully applied to discover common patterns in fact descriptions of court opinion documents \cite{drapal2023using,gray2024using}.

These applications showcase the potential of LLMs to make legal services more accessible and comprehensible to the general public. Here, we build on this important work by exploring the integration of the image modalities of LLMs. Since a lot of information relevant for self-represented litigants is contained in paper documents (e.g. letters, leases, contracts, certificates), this addition could play an important role in further increasing the usefulness of LLMs for access to justice applications.

% \subsubsection{Multi-modal models for legal tasks}
% Is there any?

\section{Experimental Design}

\begin{figure}
    \centering
    \begin{subfigure}[b]{0.45\textwidth}
        \includegraphics[width=\textwidth]{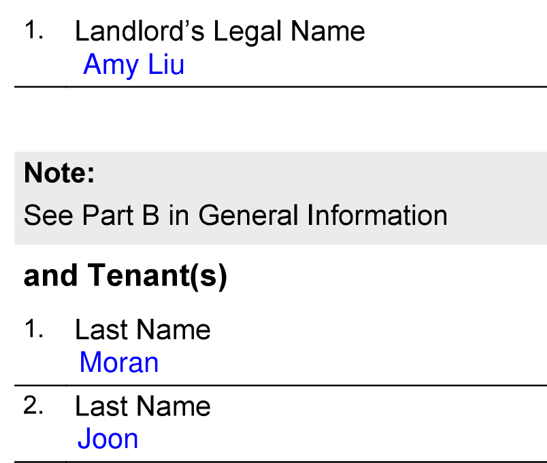}
        \caption{PDF format}
        \label{fig:formats_image1}
    \end{subfigure}
    \hfill
    \begin{subfigure}[b]{0.45\textwidth}
        \includegraphics[width=\textwidth]{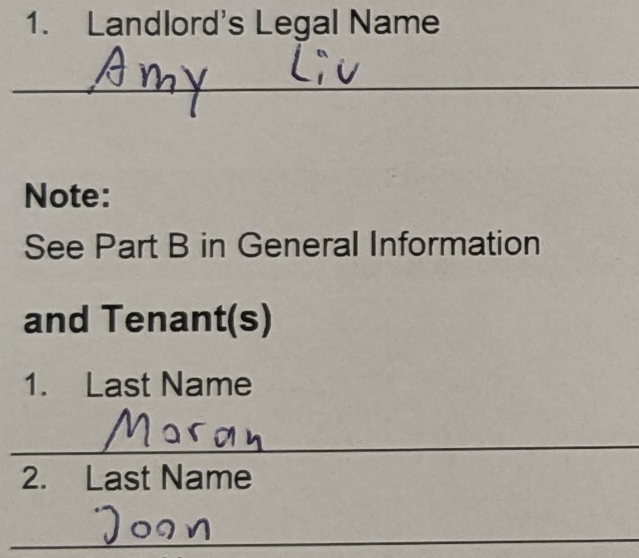}
        \caption{Paper, Neat (HD) format}
        \label{fig:formats_image2}
    \end{subfigure}
    \vskip\baselineskip
    \begin{subfigure}[b]{0.45\textwidth}
        \includegraphics[width=\textwidth]{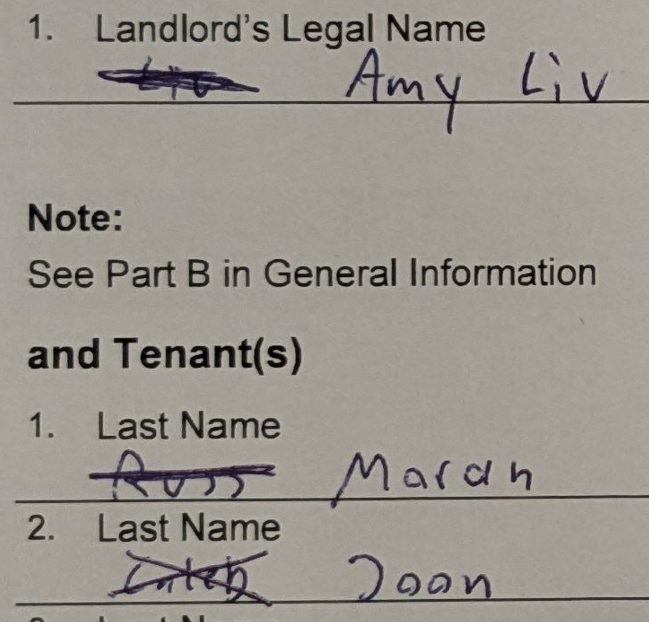}
        \caption{Paper, Sloppy (HD) format}
        \label{fig:formats_image3}
    \end{subfigure}
    \hfill
    \begin{subfigure}[b]{0.45\textwidth}
        \includegraphics[width=\textwidth]{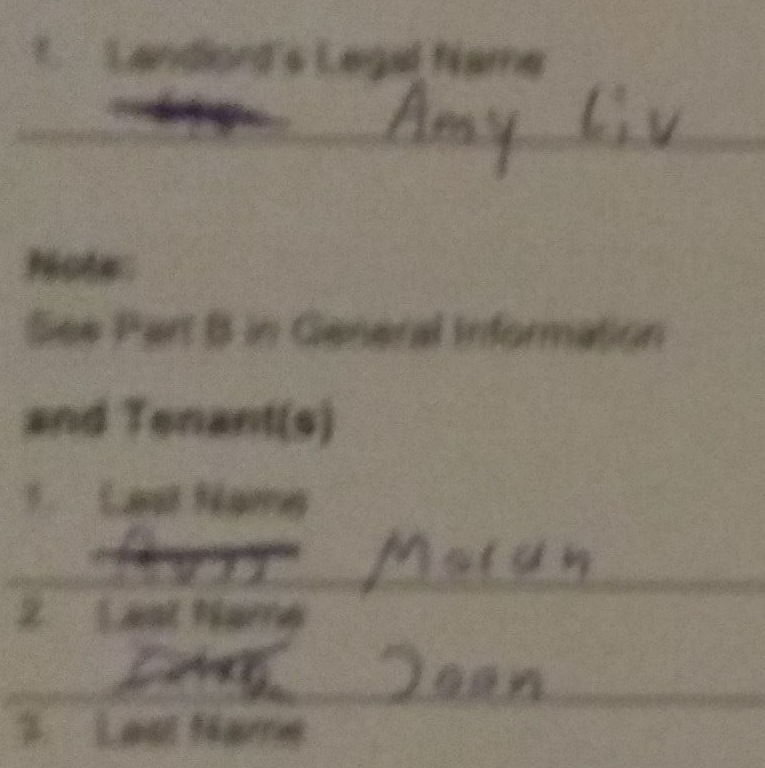}
        \caption{Paper, Sloppy (SD) format}
        \label{fig:formats_image4}
    \end{subfigure}
    \caption{Excerpts from filled-out form in the different formats used for experiments, Scenario 2.}
    \label{fig:formats}
\end{figure}

\subsection{Residential Tenancy Agreement (Standard Form of Lease)}
We created a dataset of images (photographs) of forms, based on the ``Residential Tenancy Agreement (Standard Form of Lease),'' which is the required form for entering into a lease in Ontario.\footnote{Available at \url{https://forms.mgcs.gov.on.ca/dataset/047-2229}, © King's Printer for Ontario, 2020.} This form was chosen as a representative typical form. We focused on the first page of the form, containing information about the form and fields for the landlord's name, the names of the tenants, the address of the rental unit and whether the rental unit is a condominium or not. If a user wants to file a claim regarding, e.g., deficiencies in their apartment, they are likely to require this information. Normally, they would thus need to read the form, identify the relevant information and use it for the next steps, such as typing it into a computer system or manually filling out a relevant claims form.

Next, we set up an experiment to investigate how well an LLM can extract the information contained in this form from an image of the paper form or from a screenshot of its digital version. The code and data used to perform these experiments is available online.\footnote{Available at \url{https://github.com/hwestermann/AI4A2J_analyzing_images_of_legal_documents}} 

\subsection{Dataset}
We explored three different scenarios, referring to a specific set of information filled into a form. The scenarios increase in expected task difficulty:
\begin{itemize}
    \item[S1] Scenario 1 contains common English names (Robert Edwards, Michelle Allen) and all fields are filled in.
    \item[S2] Scenario 2 contains less common names (Amy Liu, Russell Moran, Caleb Joon), two tenants, and one missing piece of information.
    \item[S3] Scenario 3 contains less common names that look similar to common names (Jame Wane, Ishan Douglasson), a missing first name for one of the tenants and various other missing fields.
\end{itemize}

\noindent Each scenario has 14 different fields that the model was expected to extract, including the first and last name of the landlord and tenant(s) and details about the rental property address. For each of the scenarios, we created various images (see Figure \ref{fig:formats} for examples). The variants are, in order of increasing expected difficulty:

\begin{enumerate}
    \item \textbf{Typed (HD)} -- The data was entered into the PDF as typed text. Then, a python framework was used to create a high-quality screenshot of the PDF (see Figure \ref{fig:formats_image1}).
    \item \textbf{Neat (HD)} -- The information was filled into the printed form in a relatively neat fashion (see Figure \ref{fig:formats_image2}). Then, an image was taken of the form using a recent phone\footnote{Google Pixel 6a, mid-range phone from 2022} in well-lit conditions.
    \item \textbf{Sloppy (HD)} -- The information was filled into the printed form in a sloppy fashion (see Figure \ref{fig:formats_image3}). For example, the text was not in the expected location and cut across the lines, there was crossed out information, and one form was filled in cursive. Then, an image was taken with the recent phone in well-lit conditions.
    \item \textbf{Neat (SD)} -- The neat version of the form described above was photographed using an older phone,\footnote{Redmi Note 4, budget phone from 2017} in poor lightning conditions.
    \item \textbf{Sloppy (SD)} -- The sloppy version of the filled out form was photographed using the older phone in poor lightning conditions (see Figure \ref{fig:formats_image4}).
\end{enumerate}

\noindent Overall, there are 3 scenarios and 5 variants of each in our experimental dataset, bringing the total amount of data to 15 samples, each with 14 possible fields. The scenarios and formats were chosen to simulate real-world situations, where users may not have access to high-quality camera equipment, and forms may not be filled out in the expected ways.

\subsection{Model \& Prompt}
To investigate the capabilities of an LLM to extract structured data from the form, we set up a pipeline that loads the images of the form and submits them to an LLM together with instructions to extract the various fields (see Figure \ref{fig:pipeline}). We used the GPT-4o model (\texttt{gpt-4o-2024-08-06}) which represents a standard model choice at the state of the art of multi-modal LLMs. We interacted with the model via the API provided by OpenAI, and set the \texttt{temperature} to 0, to minimize the chance of hallucinations as well as to make our experiments reproducible.

We provided the model with a system prompt: \textit{Analyze the provided image. Extract the values exactly as they appear, and return them in the json format specified below. If the value is missing, set that element to the string ``-''.} We also gave the model a list of the fields we expected it to extract, and an explanation of each field. For example: \textit{'landlord\_first\_name': 'The first name of the landlord, text'}. For certain fields, such as rental\_unit\_postal\_code and rental\_unit\_condominium, we specified the expected format of the return values (e.g. \textit{text, A1A 1A1} or \textit{text, yes or no}). For ease of analysis, we asked the model to provide every field as a textual value.

We then provided the images of the forms as a \textit{user message} to the model, as a base64-encoded image, according to the documentation provided by OpenAI.\footnote{\url{https://platform.openai.com/docs/guides/vision}} We did not crop or edit the images before sending them except rotating images that were sideways.

\subsection{Experiment}
\begin{figure}
    \centering
    \includegraphics[width=0.75\textwidth]{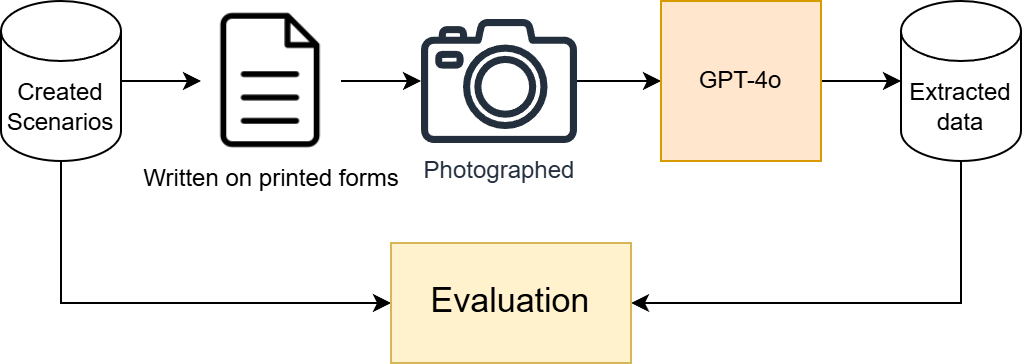}
    \caption{Diagram showing the experimental design.}
    \label{fig:pipeline}
\end{figure}

To investigate whether multi-modal LLMs can extract structured data from images of forms (RQ1), we ran the pipeline described above and captured the responses returned by the model for each sample. Then, we compared the returned data to the gold standard labels that were used to create the forms. We only considered the model to be successful if the returned text exactly matched the gold standard data, except with regards to capitalization where we allowed for flexibility. Figure \ref{fig:pipeline} shows an overview of the experimental process.

To explore the impact of the various image imperfections introduced by our experimental design (RQ2), such as complicated scenarios, missing values and poor image quality, we explored how the accuracy of the model differs between the different scenarios and sample versions.

\section{Results \& Discussion}
\begin{table}
\centering
\begin{tabular}{lcccc}
Field & S1 & S2 & S3 & Average \\
\hline
landlord\_first\_name & 1.00 & 1.00 & 0.40 & 0.80 \\
landlord\_last\_name & 1.00 & 0.40 & 0.20 & 0.53 \\
tenant\_1\_first\_name & 1.00 & 0.80 & 0.20 & 0.67 \\
tenant\_1\_last\_name & 0.80 & 0.40 & 0.40 & 0.53 \\
tenant\_2\_first\_name & 1.00 & 1.00 & 0.80 & 0.93 \\
tenant\_2\_last\_name & 1.00 & 0.40 & 0.40 & 0.60 \\
rental\_unit\_unit & 0.80 & 1.00 & 1.00 & 0.93 \\
rental\_unit\_street\_number & 0.40 & 0.20 & 0.20 & 0.27 \\
rental\_unit\_street\_name & 1.00 & 1.00 & 0.80 & 0.93 \\
rental\_unit\_city\_town & 1.00 & 1.00 & 1.00 & 1.00 \\
rental\_unit\_province & 1.00 & 1.00 & 1.00 & 1.00 \\
rental\_unit\_postal\_code & 0.40 & 0.40 & 1.00 & 0.60 \\
rental\_number\_vehicle\_spaces & 1.00 & 1.00 & 0.80 & 0.93 \\
rental\_unit\_condominium & 1.00 & 0.40 & 0.00 & 0.47 \\
\textbf{Average} & 0.89 & 0.71 & 0.59 & 0.73 \\
\end{tabular}
\caption{Correctness of extracted fields by scenario}
\label{tab:correctness}
\end{table}

\begin{table}
\centering
\begin{tabular}{lcccccc}
Field & typed, hd & neat, hd & neat, sd & sloppy, hd & sloppy, sd & Average \\
\hline
landlord\_first\_name & 1.00 & 0.67 & 1.00 & 0.67 & 0.67 & 0.80 \\
landlord\_last\_name & 1.00 & 0.67 & 0.33 & 0.33 & 0.33 & 0.53 \\
tenant\_1\_first\_name & 1.00 & 0.67 & 0.67 & 0.67 & 0.33 & 0.67 \\
tenant\_1\_last\_name & 1.00 & 0.33 & 0.67 & 0.00 & 0.67 & 0.53 \\
tenant\_2\_first\_name & 1.00 & 1.00 & 0.67 & 1.00 & 1.00 & 0.93 \\
tenant\_2\_last\_name & 1.00 & 0.67 & 0.33 & 0.67 & 0.33 & 0.60 \\
rental\_unit\_unit & 1.00 & 1.00 & 1.00 & 1.00 & 0.67 & 0.93 \\
rental\_unit\_street\_number & 1.00 & 0.33 & 0.00 & 0.00 & 0.00 & 0.27 \\
rental\_unit\_street\_name & 1.00 & 1.00 & 1.00 & 1.00 & 0.67 & 0.93 \\
rental\_unit\_city\_town & 1.00 & 1.00 & 1.00 & 1.00 & 1.00 & 1.00 \\
rental\_unit\_province & 1.00 & 1.00 & 1.00 & 1.00 & 1.00 & 1.00 \\
rental\_unit\_postal\_code & 1.00 & 0.33 & 0.67 & 0.67 & 0.33 & 0.60 \\
rental\_number\_vehicle\_spaces & 1.00 & 1.00 & 1.00 & 0.67 & 1.00 & 0.93 \\
rental\_unit\_condominium & 0.67 & 0.67 & 0.33 & 0.33 & 0.33 & 0.47 \\
\textbf{Average} & 0.98 & 0.74 & 0.69 & 0.64 & 0.60 & 0.73 \\
\end{tabular}
\caption{Correctness of extracted fields by format}
\label{tab:correctness_format}
\end{table}

\begin{figure}
    \centering
    \includegraphics[width=0.75\textwidth]{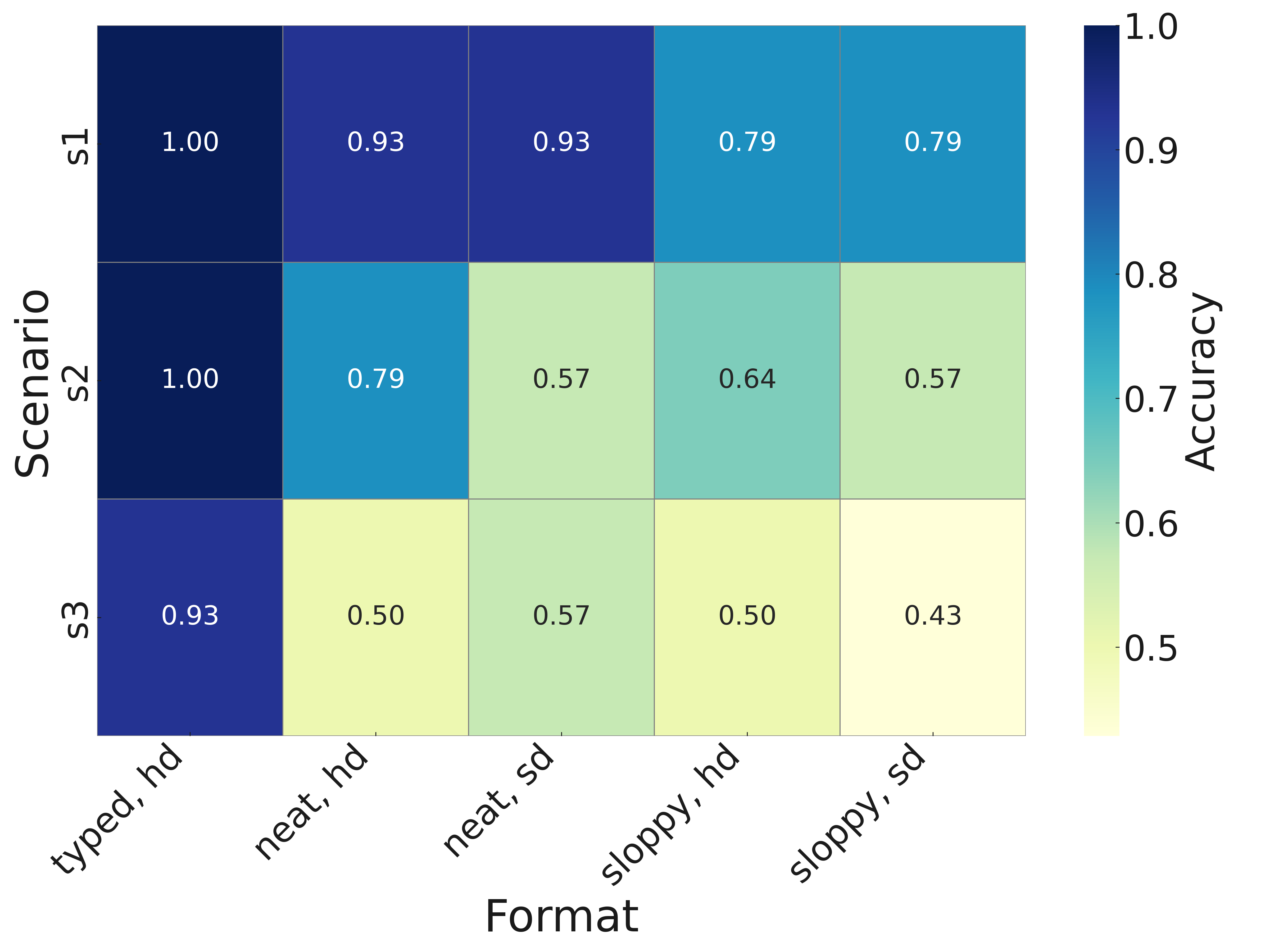}
    \caption{Accuracy heatmap for different scenarios and formats.}
    \label{fig:accuracy_heatmap}
\end{figure}

\subsection{General Results}
Table \ref{tab:correctness} shows the overall accuracy of the data extraction, segmented by field name and scenario. As we can see, on average across the formats and scenarios, 73\% of the fields were correctly extracted. Table \ref{tab:correctness_format} shows how the accuracy varies with the format used. As we can see, our assumptions with regard to the difficulty of the scenarios held up---in S1, 89\% of the fields were correctly extracted, while 71\% of the fields in S2 were correct, and only 59\% of the S3's fields were correct. Figure \ref{fig:accuracy_heatmap} shows the accuracy across the formats and scenarios. 

Looking at the results, for the values where the model was incorrect, it always picked up at least some matching letters or numbers (see e.g. Figure \ref{fig:street_number}). This indicates that even though the model sometimes struggled to extract the correct values from the fields, it had no trouble locating the information on the page---in itself a challenging task, especially considering the quality of the sloppy SD versions of the dataset, where even the form labels can be hard to read (see e.g. Figure \ref{fig:formats} and Figure \ref{fig:correct_sloppy}).

%Let us explore how various factors affected the accuracy of the extraction.

\subsection{Differences between fields}
The difficulty of accurately capturing the data varied substantially between the different fields. Certain fields (such as town and province) were perfectly captured across the formats. Looking at the data, this is likely due to the fact that the province is predefined and mentioned in multiple places on the form. The cities were picked from real cities in Ontario (Toronto, London, Hamilton). The perfect accuracy here may indicate that the model is able to benefit from its pre-training to correctly ``guess'' the value even if the image quality is lacking. Street names, which were likewise picked to be realistic (Beverly Street, New York Street, Hutchinson) were likewise extracted almost perfectly.

\begin{figure}[htbp]
    \centering
    \begin{subfigure}[b]{0.22\textwidth}
        \includegraphics[width=\textwidth]{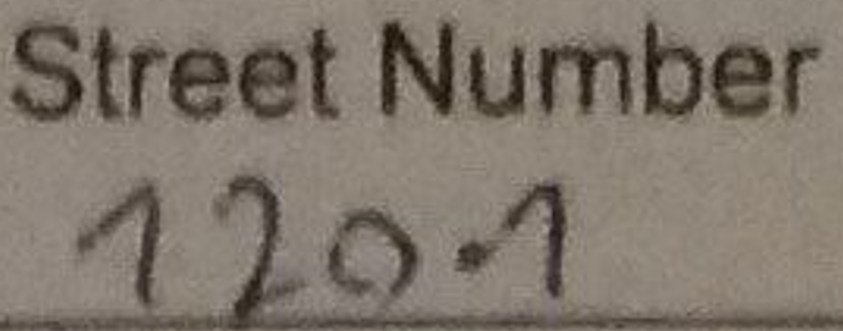}
        \caption{T: 1201, P: 1201 \\ (S1, Neat, HD)}
        \label{fig:image1}
    \end{subfigure}
    \hfill
    \begin{subfigure}[b]{0.22\textwidth}
        \includegraphics[width=\textwidth]{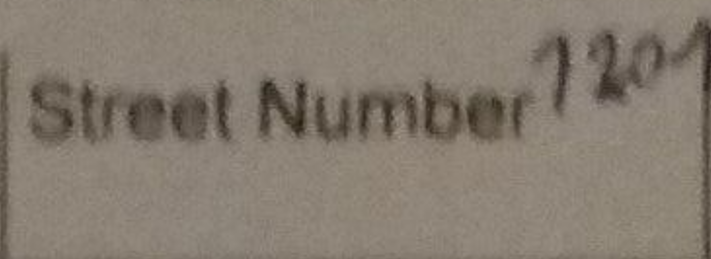}
        \caption{T: 1201, P: 120 \\ (S1, Sloppy, SD)}
        \label{fig:image2}
    \end{subfigure}
    \hfill
    \begin{subfigure}[b]{0.22\textwidth}
        \includegraphics[width=\textwidth]{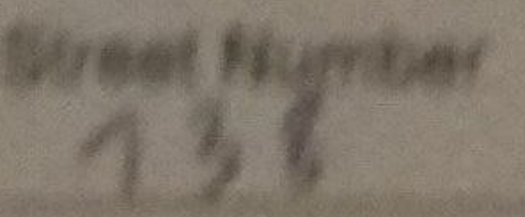}
        \caption{T: 138, P: 738 \\ (S2, Sloppy, SD)}
        \label{fig:image3}
    \end{subfigure}
    \hfill
    \begin{subfigure}[b]{0.22\textwidth}
        \includegraphics[width=\textwidth]{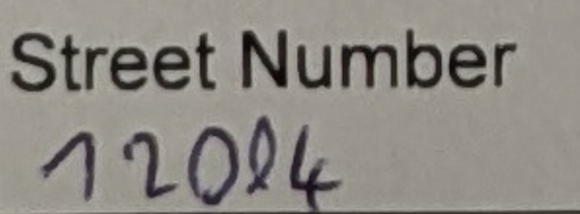}
        \caption{T: 12004, P: 12094 \\ (S3, Sloppy, HD)}
        \label{fig:image4}
    \end{subfigure}
    \caption{Excerpts from data for street numbers. T referst to the target value, while P shows the value extracted by the model.}
    \label{fig:street_number}
\end{figure}

Other values were less likely to be extracted correctly. For example, street number was only correct 27\% of the time. Figure \ref{fig:street_number} shows some of the data. As we can see, even the neat version can be difficult to read.

Another tricky field appears to be names. Here, we can see the difference between common names (such as the ones used in S1) and less common names (such as S3). Looking at the results, the model often ``corrected'' uncommon names to more common ones (Jame $\rightarrow$ Jane, Wane $\rightarrow$ Wayne). The misreadings were not always obvious---for example, ``Farmer'' sometimes became ``Sampath'' and ``Douglasson'' became ``Donaldson.'' This behavior may be unexpected, showing that LLMs may exhibit somewhat different behaviors from traditional OCR approaches.

\subsection{Differences in format}
\begin{figure}[htbp]
    \centering
    \begin{subfigure}[b]{0.31\textwidth}
        \includegraphics[width=\textwidth]{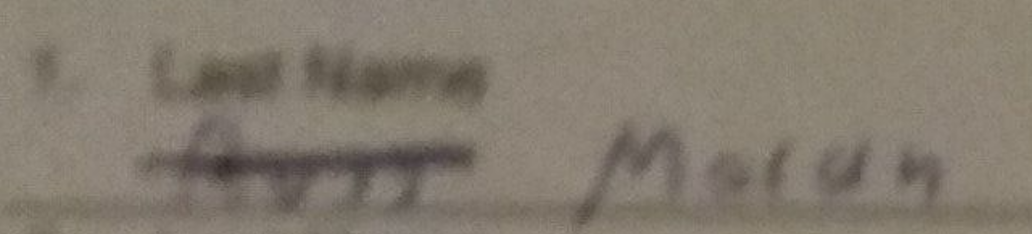}
        \caption{T: Moran, P: Moran \\ (S2, Sloppy, SD)}
        \label{fig_c:image1}
    \end{subfigure}
    \hfill
    \begin{subfigure}[b]{0.31\textwidth}
        \includegraphics[width=\textwidth]{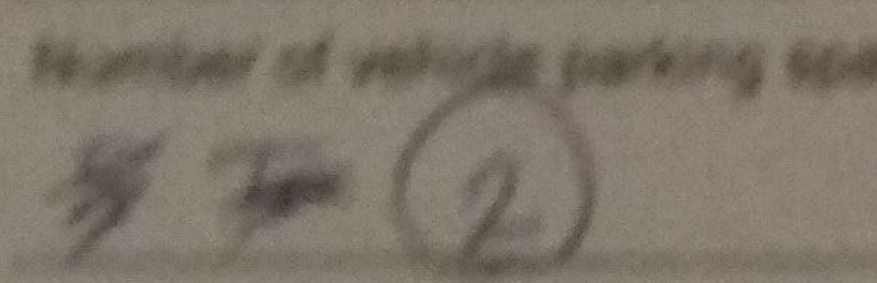}
        \caption{T: 2, P: 2 \\ (S2, Sloppy, SD)}
        \label{fig_c:image2}
    \end{subfigure}
    \hfill
    \begin{subfigure}[b]{0.31\textwidth}
        \includegraphics[width=\textwidth]{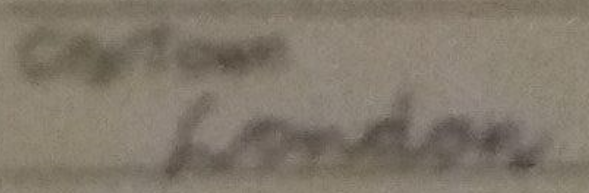}
        \caption{T: London, P: London \\ (S3, Sloppy, SD)}
        \label{fig_c:image3}
    \end{subfigure}
    \caption{Excerpts from selected samples where the model was able to correctly extract the informationd despite poor data quality. T refers to the targer value, while P refers to the prediction of the model.}
    \label{fig:correct_sloppy}
\end{figure}

Table \ref{tab:correctness_format} and Figure \ref{fig:accuracy_heatmap} show the impact of the format used. As expected, the typed PDF versions were the most straightforward to process, with the model receiving almost perfect scores. This is an impressive result, and shows that such models could already be useful for analyzing electronic documents. The quality of the results decreased when working with the printed versions. For the neat written form for S1, they still come out relatively strong. However, in the other formats and the sloppy forms, the accuracy decreased notably. Considering the quality in Figure \ref{fig:formats}, this is not unexpected. Figure \ref{fig:correct_sloppy} shows some selected examples where the model was able to correctly locate and extract a value despite poor data quality.

\subsection{Analysis}
Overall, it appears there is considerable potential for the use of multi-modal LLMs to extract data from images of forms. It seems like the models have no trouble locating the correct field in the forms, and the extracted results are relatively accurate, especially for the high-quality formats and for fields where the model could benefit from information it already knows, such as a city in a specific province. The fact that the model is able to take an image of a form filled in with a pen in handwriting, and obtain reasonable results, is overall an impressive achievement. This is especially the case since multi-modal models are much more recent than the purely text-based LLMs, meaning that we may be able to expect further performance improvements in the coming months. Especially in collaboration with a human that can verify the data captured by the model and correct mistakes, one can imagine multiple applications of such technology (e.g., allowing a user to take an image of forms, legal documents or letters, automatically extracting the relevant fields and then filling out new forms or providing relevant legal information). 

At the same time, it is important to remain cautious. For example, the results clearly demonstrate the importance of image quality when extracting the data. If this model is provided to the public, one has to remain aware of the implications thereof for the digital divide, as a modern phone and good lightning conditions are important to get good results, which may exclude certain groups from using such tools. At the same time, the prices of phones and the quality of the cameras are constantly improving, meaning that these barriers may come down over time.

Another issue lies in the preference of the models for more common data. Names such as Robert and Michelle were easier for the model to spot than, e.g., Jame or Joon. This shows that the underlying token distributions embedded in the models can have an impact on their performance, and may risk introducing biases in society if not properly evaluated.

\section{Conclusion \& Future Work}
We investigated the use of multi-modal LLMs to extract structured information from images of forms. The results are promising, with implications for access to justice. Such models could be a powerful way to support laypeople and self-represented litigants, by extracting and analyzing crucial data contained in printed documents.

To further develop these capabilities, we are planning to conduct more large-scale studies with more varied data, to adjust the prompt and the model used to investigate the performance impacts thereof, and to see how the functionality described here can be integrated into a system to provide legal information or fill in forms.

%% Define the bibliography file to be used
\bibliography{sample-ceur}

\end{document}